\documentclass[letterpaper,10pt,conference]{ieeeconf}
\IEEEoverridecommandlockouts
\usepackage{times}
\usepackage{cite}
\newcommand{\citep}{\cite}
\usepackage{url}
\providecommand{\href}[2]{#2}
\usepackage{graphicx,float,capt-of}
\usepackage[caption=false]{subfig}
\usepackage{amsmath,amssymb}
\usepackage{booktabs,tabularx,array,multirow,hhline}
\newcommand{\tablebodyfont}{\fontsize{8.1pt}{9pt}\selectfont}
\newcommand{\tablebodystyle}{%
  \tablebodyfont
  \renewcommand{\arraystretch}{1.18}%
  \setlength{\arrayrulewidth}{0.675pt}%
}
\usepackage[skins,breakable]{tcolorbox}
\let\labelindent\relax
\usepackage{enumitem,longtable,adjustbox}
\usepackage{pgfplots}
\pgfplotsset{compat=1.18}
\usepackage[table]{xcolor}
\makeatletter
\providecommand{\insert@pcolumn}{\insert@column}
\makeatother
\usepackage{algorithm,algpseudocode}
\usepackage{placeins}
\usepackage{newunicodechar}
\DeclareUnicodeCharacter{1D7CE}{0}
\DeclareUnicodeCharacter{1D7CF}{1}

\begin{document}

\title{\LARGE \bf Veo-Act: Enhancing VLA Policies with Frontier Video Models}

\author{\normalsize
Zhongru Zhang$^{1,*}$, Chenghan Yang$^{1,*}$, Qingzhou Lu$^{1,*}$, Yanjiang Guo$^{1,*,\dagger}$\\
\normalsize Jianke Zhang$^{1}$, Yucheng Hu$^{1}$, Jianyu Chen$^{1}$\\
\normalsize $^{1}$Tsinghua University\quad $^{*}$Equal Contribution\quad $^{\dagger}$Project Lead}

\IEEEaftertitletext{\vspace{-12pt}}
\maketitle
\thispagestyle{empty}
\pagestyle{empty}
\raggedbottom

\begin{abstract}
Video generation models can produce coherent visual sequences depicting object motion and interactions. We investigate how frontier video generation models can complement vision-language-action policies to enhance generalizable robotic manipulation. VLA policies have become a dominant paradigm for robot learning, but their action-oriented adaptation of pretrained VLMs can weaken semantic generalization, limiting robustness in ambiguous or out-of-distribution manipulation scenarios. We use video models as visual planners, motivated by their potential to generalize across complex scenes and their priors over hand motion. However, manipulation methods based on video models often lack the precision and temporal responsiveness needed for low-level dexterous interaction.

To address this gap, we present \textit{Veo-Act}, a hierarchical framework with Veo-3.1 as a high-level motion planner and a VLA policy as the low-level executor. A multi-head inverse dynamics model converts generated frame pairs into actions and learns an interaction gate to trigger the handoff to reactive VLA control. Experiments in simulation and on a real robot show improved instruction following and overall task success over the baseline VLA in novel and semantically complex manipulation settings, supporting the complementary roles of video planning and reactive interaction.

\end{abstract}

\section{Introduction}
\label{sec:intro}

Robots must manipulate objects with unfamiliar geometries in environments that differ from their training data. Vision-language-action (VLA) models adapt pretrained vision-language models (VLMs) for robotic control~\cite{kim2024openvla, black2024pi_0,intelligence2025pi_,guo2025improving,zhang2025up,zhang2024hirt,zhang2025unicod}.

Building strong VLA models remains difficult. Large-scale real-world robot data collection is costly and time-consuming~\cite{brohan2022rt}, leaving existing manipulation datasets limited in task types, object configurations, camera viewpoints, and interaction patterns. Consequently, VLA policies can overfit to their training environments and generalize poorly to unfamiliar objects and scenes. Knowledge transfer from VLMs to VLAs is also challenging: even with web-scale co-training, downstream VLA policies still cannot fully preserve the generalization capability of pretrained VLMs~\cite{driess2025knowledge, hancock2025actions,zhang2026vlm4vla}.

\begin{figure}[t]
  \centering
  \includegraphics[width=0.915200\columnwidth]{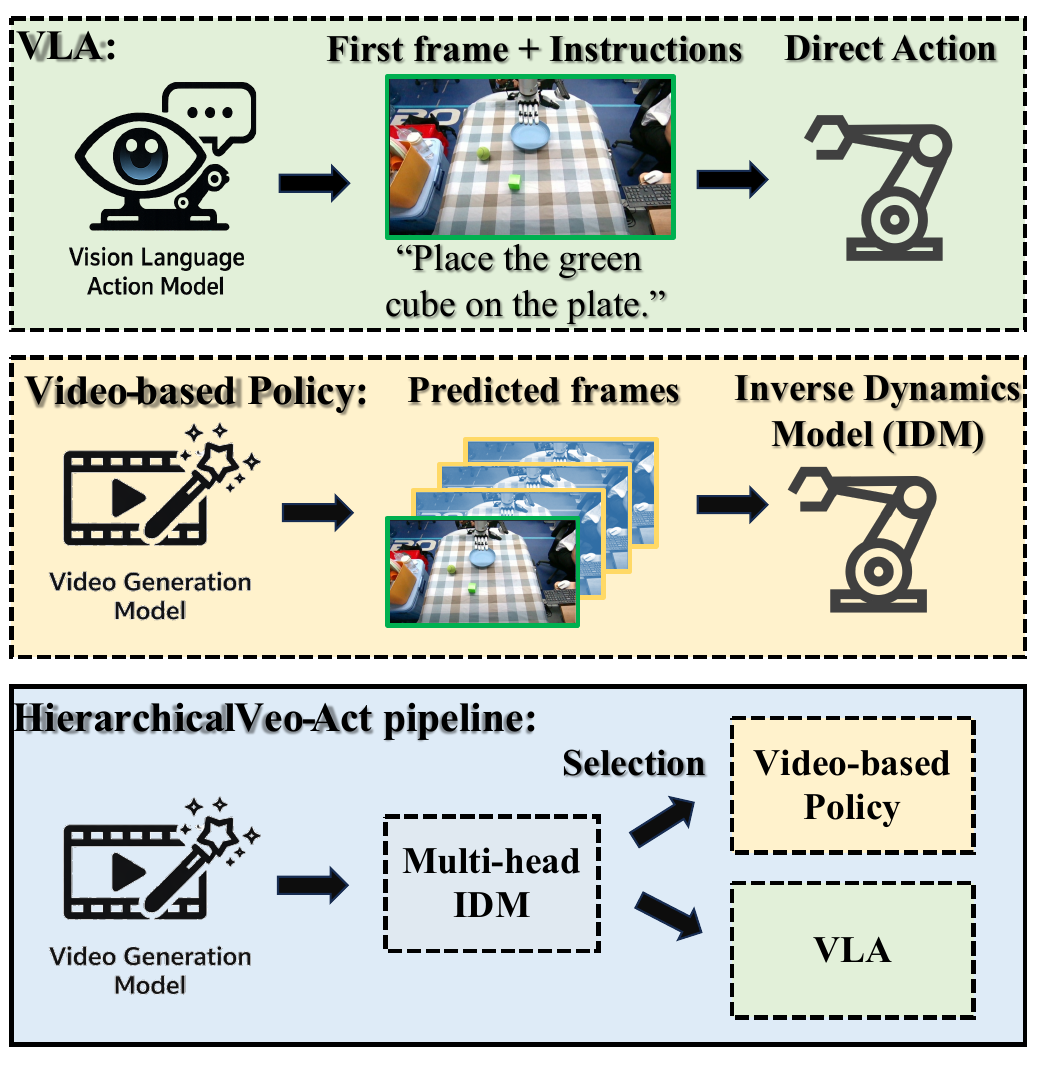}
  \caption{Comparison of three control pipelines. (i) A VLA adapts a VLM by introducing an action modality, but this adaptation can weaken semantic generalization. (ii) ``Video model + IDM'' preserves video model priors and is effective for semantic motion planning, but recovered actions can lose accuracy during contact-rich interaction. (iii) Our \textbf{Veo-Act} is a hierarchical pipeline that automatically switches between a video planner and a VLA, leveraging the strengths of both models.}
  \label{fig:intro_teaser}
\end{figure}

In parallel, video generation models have rapidly advanced, producing realistic frames and showing strong physical dynamics modeling and semantic understanding~\cite{ho2022imagen, singer2022make, kondratyuk2023videopoet, brooks2024video}. Recent works use generated visual plans or video model representations for robot control~\cite{hu2024video, kim2026cosmos, ye2026world, shen2025videovla}. These methods demonstrate generalization in robotic manipulation, but most rely on video models weaker than frontier foundation video models~\cite{deepmindveo2024, openai2024sora}; moreover, video generation remains inaccurate in contact-rich interactions, where VLA policies often provide more precise and responsive low-level control~\cite{guo2025ctrl}. This raises the question: \textit{how can video generation models complement VLA policies to improve generalization in robotic manipulation?}

We observe that, even in complex dexterous manipulation scenes, Veo-3.1-fast often generates approximately correct task-level trajectories and faithfully follows task instructions~\cite{deepmindveo2024}. Although prior work demonstrates action recovery from generated videos or visual predictions under zero-demonstration settings~\cite{hu2024video, liang2025video, li2025novaflow}, our results reveal a substantial gap between recovered actions and executable robot trajectories, especially in contact-rich dexterous interactions.

The potential to use video models' generalization across complex scenes and priors over hand motion motivates our choice of a visual planner. We therefore propose a hierarchical manipulation framework: Veo-3.1-fast videos provide high-level motion plans, a VLA provides reactive low-level manipulation control, and a multi-head inverse dynamics model (IDM) predicts actions and an interaction indicator for switching between video-based planning and VLA-based execution. In experiments, the integrated Veo-Act pipeline improves the pooled overall task success rate of a strong VLA baseline, \(\pi_{0.5}\)~\cite{intelligence2025pi_}, from 45\% to 80\% on simulated and real-world dexterous-hand platforms.

\textbf{In summary, our findings and contributions include:}\par\nopagebreak[4]
\noindent (1) We verify that Veo-3.1-fast can provide semantically aligned and temporally coherent task-level visual plans for dexterous manipulation, while actions recovered from such videos remain insufficient for reliable low-level contact-rich execution. \\
\noindent (2) We propose a hierarchical framework that combines video models for task-level motion planning with a VLA for reactive interaction. A multi-head IDM converts generated image transitions into a fixed action queue and predicts interaction gates from real observations to trigger the handoff between them.

\section{Related Work}
\label{sec:related_work}

\subsection{Vision-Language-Action Models}

Generalist robot learning is increasingly dominated by Vision-Language-Action (VLA) models~\cite{hu2026bagelvlaenhancinglonghorizonmanipulation, guo2026vlaw, chen2025villa}, including RT-2~\citep{zitkovich2023rt}, OpenVLA~\citep{kim2024openvla}, $\pi_0$~\citep{black2024pi_0} and $\pi_{0.5}$~\citep{intelligence2025pi_} (flow matching), and DexVLA~\citep{wen2025dexvla} (enhanced with diffusion experts). Built on pretrained vision-language models (VLMs), these methods aim to transfer semantic and physical knowledge to robotic control, with recent works further improving long-horizon manipulation through phase-aware or interleaved vision-language-action generation~\citep{fan2025longvlaunleashinglonghorizoncapability, hu2026bagelvlaenhancinglonghorizonmanipulation}.

Methods for preserving pretrained knowledge include language co-training~\citep{intelligence2025pi_} and action-as-language representations~\citep{hancock2025actions}, but fully retaining pretrained VLM generalization remains challenging, and stronger VLM backbones do not necessarily yield stronger VLA policies.

\subsection{Video-Based Policies and World Action Models}

Recent video generation models, including Sora~\citep{brooks2024video}, Seedance~\bstctlcite{SeedanceShortAuthors}\citep{seedance2026seedance20advancingvideo}\bstctlcite{RestoreFullAuthors}, Veo-3.1-fast~\citep{deepmindveo2024}, and Wan~\citep{wan2025wan}, as well as self-supervised video models including V-JEPA~2~\citep{assran2025vjepa2} and action-conditioned world models such as Unified World Models~\citep{zhu2025uwm}, are trained on large-scale video or robotic datasets and acquire physical priors such as object permanence, motion continuity, and coarse collision dynamics. These capabilities have motivated video-based robot policies beyond inverse dynamics pipelines.

One research line uses a video model for high-level visual planning, then grounds the generated future in robot actions. Imagine-then-execute methods recover actions with inverse dynamics or tracking~\citep{du2023learning, feng2025vidar, tan2025anypos, mi2026tc, yang2023unisim}. Later systems distill generated videos into object-centric 3D flow and classical control~\citep{li2025novaflow}, treat the video generator itself as a generalizable manipulator~\citep{shen2025videovla, liang2025video, kim2026cosmos}, or use large video planners for generalizable control~\citep{chen2025largevideoplannerenables, guo2024prediction}. A second line requires no pixel decoding at control time: Video Prediction Policy and FLARE condition policies on predictive video representations~\citep{hu2024video, zheng2025flarerobotlearningimplicit}, DreamGen uses video world models to improve policy generalization~\citep{jang2025dreamgen}, and Unified Video Action models jointly train video and action while skipping video decoding at test time~\citep{li2025uva}.

World Action Models (WAMs) make this coupling explicit by jointly modeling how observations evolve under action and using that dynamics prior for control~\citep{ye2026world, zhu2025uwm, cen2025worldvla}. Fast-WAM further shows that much of the benefit can come from video co-training rather than test-time future imagination, using a Wan backbone while predicting actions in a single forward pass~\citep{yuan2026fastwam}. These methods demonstrate that policies based on video models can be highly competitive with VLAs and already use strong open video backbones such as Wan.

Visually plausible prediction is nevertheless not the same as physically executable dexterous interaction. Grounding generated futures remains sensitive to video artifacts in contact-rich scenes~\cite{guo2025ctrl, deng2026rethinking, hsieh2025dexman, chen2025gendexhand}. Veo-Act addresses this gap by handing off video-guided execution to a reactive VLA during interaction.

\section{Method}
\label{sec:method}

\begin{figure*}[t]
  \centering
  \includegraphics[width=0.675\textwidth,height=0.27\textheight,keepaspectratio]{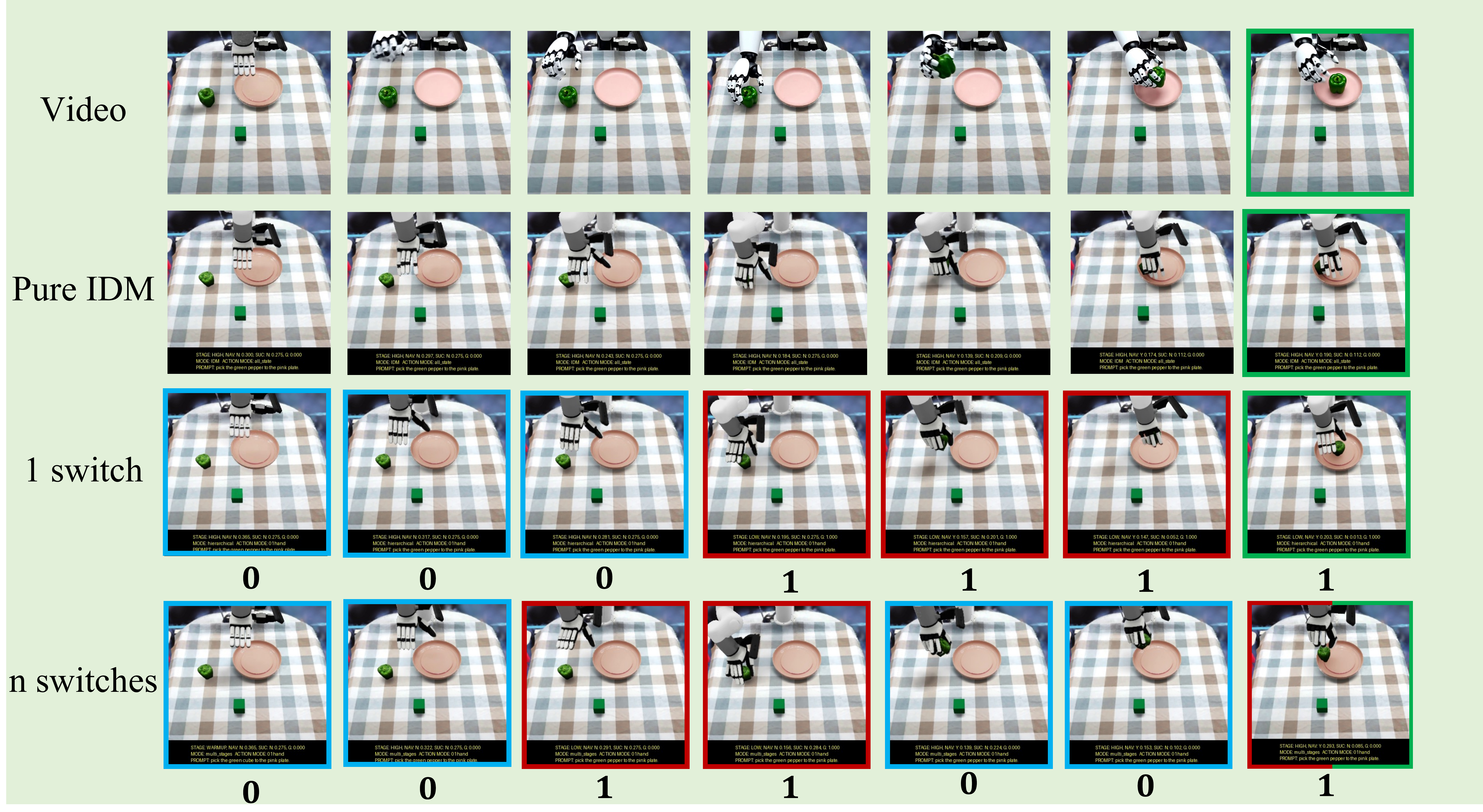}
  \caption{\textbf{Three inference paradigms.} Top: generated video trajectory ending at the intended task outcome. Rows 2--4: executed trajectories of pure IDM, Veo-Act retaining low-level control after the first switch, and multi-switch Veo-Act, respectively. Here, 0/1 denote IDM/VLA control, distinct from the training interaction labels.}
  \label{fig:traj}
\end{figure*}

Veo-Act converts a generated visual plan into an action queue and uses an interaction gate to trigger reactive VLA control. Figure~\ref{fig:traj} compares execution variants, and Fig.~\ref{fig:pipeline} shows the pipeline and notation.

\subsection{Video Generation and Multi-Head IDM}

Let \(I_t\) denote the global-view image, \(W_t\) the wrist-view image, \(s_t\) the robot state, and \(\ell\) the language instruction at real execution step \(t\). Given \(I_0\) and \(\ell\), we query a video model for a video showing how the task should be completed. We denote the generated frame sequence as \(I^{*}_{0:n}=\{I_0,I^{*}_1,\ldots,I^{*}_n\}\), with \(I^{*}_0=I_0\) and \(n\) synthesized future frames. This sequence provides a high-level motion prior in image space.

We employ a multi-head \textbf{inverse dynamics model (IDM)} that maps image transitions to robot actions and predicts a gate for switching to low-level dexterous interaction. The IDM receives only two consecutive global-view images; recorded robot configurations provide supervision, not input features. For training pair \((I_{i-1},I_i)\), \(a^{\mathrm{data}}_i\) is the configuration recorded at the later frame \(i\): absolute end-effector pose and hand joint positions, not a displacement between frames. \textbf{DINOv3}~\cite{simeoni2025dinov3} encodes each frame; the concatenated pooled features feed two MLP heads. The action head predicts \(a^{\mathrm{IDM}}_i\), while the interaction head produces a logit whose sigmoid gives \(G_i\in[0,1]\). Thus, \((a^{\mathrm{IDM}}_i,G_i)=f_{\mathrm{IDM}}(I_{i-1},I_i)\). The configuration comprises 3 end-effector position coordinates, a 6D representation of its 3-DoF orientation, and 12 hand joint coordinates, with arm poses in the robot base frame. These 21 representation dimensions correspond to a robot with 19 joint DoFs (7 arm, 12 hand); inverse kinematics maps the 6-DoF end-effector pose to the redundant 7-DoF arm. The interaction gate is a separate output.

Before execution, the IDM processes each generated frame pair to obtain \((a^{*}_j,G^{*}_j)=f_{\mathrm{IDM}}(I^{*}_j,I^{*}_{j+1})\), for \mbox{\(j=0,\ldots,n-1\)}. Here, \(j\) indexes planned transitions, distinct from the real execution step \(t\); \(a^{*}_j\) estimates the configuration depicted by the later frame \(I^{*}_{j+1}\). We jointly smooth concatenated actions and gate probabilities into paired entries \((\hat a^{*}_k,\hat G^{*}_k)\), \(k=0,\ldots,m\). The smoother filters each channel and repeats complete entries at selected keypoints to extend the trajectory. Actions and gates therefore retain the same row indices after smoothing and temporal extension. We execute the fixed queue sequentially and update switching decisions online from real observations, without regenerating the remaining video plan at each step.

\begin{figure*}[t]
  \centering
  \vspace*{5pt}
  \includegraphics[width=0.8\textwidth]{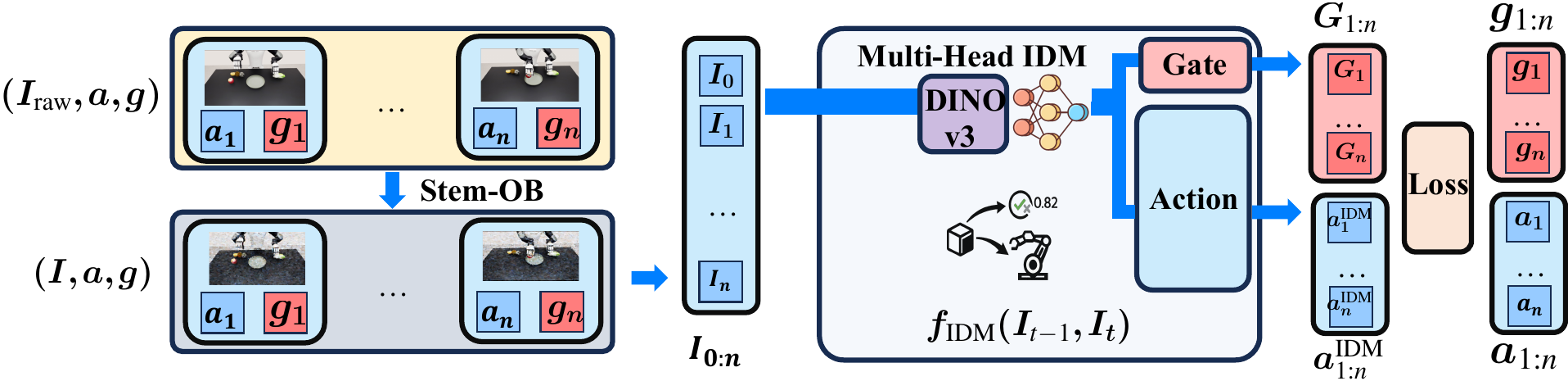}
  \caption{\textbf{Multi-head IDM training pipeline.} STEM-OB augments consecutive global-view images before they enter the trainable DINOv3 encoders. The recorded configurations $a_{1:n}$ (shorthand for the training targets $a^{\mathrm{data}}_{1:n}$) and interaction labels $g_{1:n}$ are supervision targets, not model inputs. The heads jointly learn action regression and interaction detection; samples without interaction labels contribute only the action loss.}
  \label{fig:train}
  \vspace{-5pt}
\end{figure*}

We train the multi-head IDM end-to-end with unfrozen DINOv3 encoders; both output losses update the visual features shared by the two heads. Only the IDM undergoes this joint optimization; the video generator and VLA are not jointly trained. The action and interaction heads use Huber loss and binary cross-entropy, respectively. For training sample \(i\), the total loss is \(\mathcal{L}=\lambda_{\mathrm{act}}\mathcal{L}_{\mathrm{act}}(a^{\mathrm{data}}_i,a^{\mathrm{IDM}}_i)+\lambda_{\mathrm{gate}}\mathcal{L}_{\mathrm{gate}}(g_i,G_i)\). The label \(g_i\in\{0,1\}\) denotes non-interaction or interaction, and the coefficients balance the two losses. Only interaction-labeled samples receive the gate loss; additional action-only samples supervise the action head. Figure~\ref{fig:train} shows the training pipeline.

\subsection{Hierarchical Planning and Execution}

\begin{figure*}[t]
  \centering
  \vspace*{1pt}
  \includegraphics[width=0.823680\textwidth]{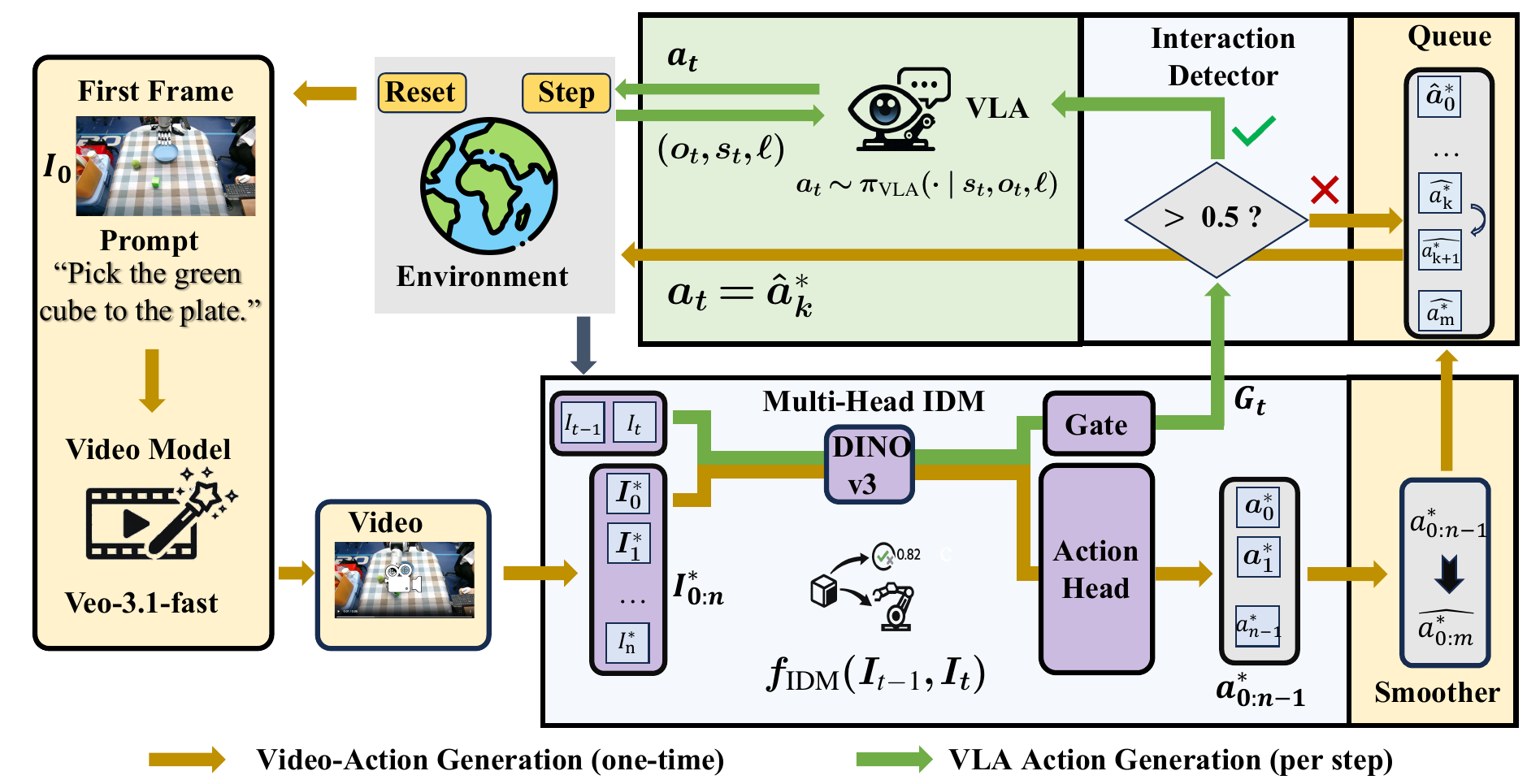}
  \caption{\textbf{Hierarchical planning and control pipeline.} A video model generates $I^{*}_{0:n}$ from $I_0$ and language instruction $\ell$. The image-only IDM predicts action--gate pairs, jointly smoothed into a fixed queue indexed by $k=0,\ldots,m$. Online gate $G_t$ uses consecutive real global-view images; the VLA receives robot state $s_t$, both views $o_t=(I_t,W_t)$, and $\ell$. The general controller can resume the remaining queue after interaction; main experiments use one VLA handoff.}
  \label{fig:pipeline}
  \vspace{-1pt}
\end{figure*}

After video generation and action conversion, the robot executes step by step. We reserve $a_t$ for the configuration command issued at execution step $t$, distinct from the recorded training target $a^{\mathrm{data}}_i$; the command need not match the configuration actually reached at that step. Cursor $k$, initially zero, indexes the next queue entry; $t$ indexes real execution steps. The remaining queue is $\mathcal{Q}_t=\{(\hat a^{*}_r,\hat G^{*}_r)\}_{r=k}^{m}$. In the video-guided stage, the controller executes $a_t=\hat a^{*}_k$ and increments $k$; the cursor pauses during VLA control.

In parallel, the IDM interaction head computes $G_t$ from consecutive real images $(I_{t-1},I_t)$. If $G_t$ exceeds threshold $\tau$ (e.g., 0.5) for a short persistence window, control switches to the low-level policy for target interaction; otherwise, the controller continues consuming the planned queue.

When enabled, the low-level policy receives both camera views $o_t=(I_t,W_t)$, robot state $s_t$, and the same task instruction $\ell$, and outputs $a_t\sim\pi_{\mathrm{VLA}}(\cdot\mid s_t,o_t,\ell)$. The baseline VLA uses the same two-view observation interface. The general multi-switch controller continues evaluating $G_t$ online and returns control to the queue after $G_t$ stays low for the required persistence window. Starting at the paused cursor, we skip consecutive entries whose stored $\hat G^{*}_k$ exceeds the queue gate threshold and resume at the first low-gate entry. This uses gates already aligned with planned actions, without matching real and generated timestamps. If no entry remains, the controller holds the current configuration. In the single-switch configuration used for the main results, the VLA remains active after the first handoff.

\section{Experiments}
\label{sec:exp}

We evaluate Veo-Act in three confounded simulation settings and three real robot settings designed to expose a VLA baseline's failures in following instructions.

\subsection{Experimental Setup}
\label{sec:setup}

\subsubsection{Real Robot and Simulation Environments}
We use a 7-DoF arm with a 12-DoF dexterous hand and two RGB cameras: one covering the workspace and one wrist-mounted for close-up views. Video generation and IDM prediction use only global images; both VLAs receive both views, robot state, and task instruction, with Veo-Act activating its VLA after handoff. For large-scale data collection and realistic evaluation, a high-fidelity IsaacLab simulation~\cite{makoviychuk2021isaac, mittal2025isaac, NVIDIA_Isaac_Sim} mirrors the physical setup.

\subsubsection{Dataset}
We train the multi-head IDM on 300k simulation frame pairs from 100--200-step trajectories interleaving random robot motions and object interactions. At each step, we record the global image and corresponding 21-dimensional single-arm configuration target $a^{\mathrm{data}}_i$ (Sec.~\ref{sec:method}); this target supervises actions and is not an IDM input. Gate labels are 1 for interaction and 0 for non-interaction, including navigation and transport after interaction completion.

For robustness, we add 100k purely random-motion simulation samples and 150k real-world samples from the physical platform for action prediction only, strengthening visual representations and reducing the sim-to-real gap. STEM-OB~\citep{hu2024stem} observation-level noise augmentation of all collected trajectories further improves cross-domain generalization.

\subsubsection{Evaluation Setup}
We observe that dexterous VLA policies often confuse semantically similar objects, over-rely on wrist-camera visibility, and are sensitive to object placement, reducing robustness under distribution shifts. Our simulated and real robot object-placement settings expose generalization differences by inducing semantic or perceptual failures in the VLA baseline.
The OOD evaluation uses objects, scene configurations, and tasks absent from the IDM and VLA robot training data. For each simulation setting, we compare the baseline and Veo-Act under Control, which removes the confounding factor, and Experimental, which includes it to stress-test generalization. Each paired Control and Experimental scene keeps the target object at the same initial position; the distractor configuration changes between the two conditions. Real robot results are reported under the Experimental condition.

\begin{table*}[t]
  \vspace*{5pt}
  \centering
  \footnotesize
  \caption{\textbf{Simulation and real robot results.} Simulation compares $\pi_{0.5}$, Veo-Act, and the video-based VPP; real robot experiments compare $\pi_{0.5}$ and Veo-Act. Bold marks the highest rate for each condition and metric, including ties.}
  \label{tab:all_results}
  \vspace{-5pt}
  \fontsize{9.8138pt}{11.0405pt}\selectfont
  \setlength{\tabcolsep}{2pt}
  \renewcommand{\arraystretch}{1.05}
  \setlength{\arrayrulewidth}{0.9200pt}
  \newcommand{\thincline}[1]{\noalign{\global\arrayrulewidth=0.4907pt}\cline{#1}\noalign{\global\arrayrulewidth=0.9200pt}}

  \begin{tabularx}{\textwidth}{ll*{12}{>{\centering\arraybackslash}X}}
    \hline
    \multicolumn{2}{c}{} & \multicolumn{12}{c}{\textbf{Simulation}} \\
    \hline
    \multicolumn{2}{c}{}
      & \multicolumn{4}{c}{\textbf{Wrist-camera invisible}}
      & \multicolumn{4}{c}{\textbf{Similar-object distractors}}
      & \multicolumn{4}{c}{\textbf{Pass-by interaction}} \\
    \thincline{3-14}
    \textbf{Method} & \textbf{Metric}
      & \multicolumn{2}{c}{\textbf{Control}} & \multicolumn{2}{c}{\textbf{Experimental}}
      & \multicolumn{2}{c}{\textbf{Control}} & \multicolumn{2}{c}{\textbf{Experimental}}
      & \multicolumn{2}{c}{\textbf{Control}} & \multicolumn{2}{c}{\textbf{Experimental}} \\
    \thincline{3-14}
    & & \textbf{Suc/All} & \textbf{Rate} & \textbf{Suc/All} & \textbf{Rate}
      & \textbf{Suc/All} & \textbf{Rate} & \textbf{Suc/All} & \textbf{Rate}
      & \textbf{Suc/All} & \textbf{Rate} & \textbf{Suc/All} & \textbf{Rate} \\
    \hline

    \rowcolor{gray!20}
    & Instr-follow
      & 28/30 & 0.93
      & 25/30 & \textbf{0.83}
      & 28/30 & 0.93
      & 28/30 & \textbf{0.93}
      & 28/30 & \textbf{0.93}
      & 15/30 & \textbf{0.50} \\
    \rowcolor{gray!20}
    \multirow{-2}{*}{Veo-Act} & Overall
      & 28/30 & 0.93
      & 20/30 & \textbf{0.67}
      & 28/30 & 0.93
      & 28/30 & \textbf{0.93}
      & 26/30 & \textbf{0.87}
      & 14/30 & \textbf{0.47} \\
    \hline

    \multirow{2}{*}{$\pi_{0.5}$} & Instr-follow
      & 30/30 & \textbf{1.00}
      & 11/30 & 0.37
      & 29/30 & \textbf{0.97}
      & 14/30 & 0.47
      & 11/30 & 0.37
      & 1/30 & 0.03 \\
    & Overall
      & 29/30 & \textbf{0.97}
      & 10/30 & 0.33
      & 29/30 & \textbf{0.97}
      & 12/30 & 0.40
      & 10/30 & 0.33
      & 0/30 & 0.00 \\
    \hline

    \multirow{2}{*}{VPP} & Instr-follow
      & 26/30 & 0.87
      & 18/30 & 0.60
      & 16/30 & 0.53
      & 8/30 & 0.27
      & 7/30 & 0.23
      & 3/30 & 0.10 \\
    & Overall
      & 23/30 & 0.77
      & 15/30 & 0.50
      & 13/30 & 0.43
      & 6/30 & 0.20
      & 5/30 & 0.17
      & 1/30 & 0.03 \\
    \hline

    \multicolumn{14}{c}{} \\
    \hline
    \multicolumn{2}{c}{} & \multicolumn{12}{c}{\textbf{Real robot}} \\
    \hline
    \multicolumn{2}{c}{}
      & \multicolumn{4}{c}{\textbf{Similar-object distractors}}
      & \multicolumn{4}{c}{\textbf{Pass-by interaction}}
      & \multicolumn{4}{c}{\textbf{Richer semantics}} \\
    \thincline{3-14}
    & & \multicolumn{4}{c}{\textbf{Experimental}}
      & \multicolumn{4}{c}{\textbf{Experimental}}
      & \multicolumn{4}{c}{\textbf{Experimental}} \\
    \thincline{3-14}
    \textbf{Method} & \textbf{Metric}
      & \multicolumn{2}{c}{\textbf{Suc/All}} & \multicolumn{2}{c}{\textbf{Rate}}
      & \multicolumn{2}{c}{\textbf{Suc/All}} & \multicolumn{2}{c}{\textbf{Rate}}
      & \multicolumn{2}{c}{\textbf{Suc/All}} & \multicolumn{2}{c}{\textbf{Rate}} \\
    \hline

    \rowcolor{gray!20}
    & Instr-follow
      & \multicolumn{2}{c}{15/16} & \multicolumn{2}{c}{\textbf{0.94}}
      & \multicolumn{2}{c}{12/13} & \multicolumn{2}{c}{\textbf{0.92}}
      & \multicolumn{2}{c}{18/19} & \multicolumn{2}{c}{\textbf{0.95}} \\
    \rowcolor{gray!20}
    \multirow{-2}{*}{Veo-Act} & Overall
      & \multicolumn{2}{c}{12/16} & \multicolumn{2}{c}{\textbf{0.75}}
      & \multicolumn{2}{c}{11/13} & \multicolumn{2}{c}{\textbf{0.85}}
      & \multicolumn{2}{c}{15/19} & \multicolumn{2}{c}{\textbf{0.79}} \\
    \hline

    \multirow{2}{*}{$\pi_{0.5}$} & Instr-follow
      & \multicolumn{2}{c}{9/16} & \multicolumn{2}{c}{0.56}
      & \multicolumn{2}{c}{3/13} & \multicolumn{2}{c}{0.23}
      & \multicolumn{2}{c}{4/19} & \multicolumn{2}{c}{0.21} \\
    & Overall
      & \multicolumn{2}{c}{8/16} & \multicolumn{2}{c}{0.50}
      & \multicolumn{2}{c}{2/13} & \multicolumn{2}{c}{0.15}
      & \multicolumn{2}{c}{2/19} & \multicolumn{2}{c}{0.11} \\
    \hline

  \end{tabularx}
\end{table*}

\begin{figure}[!htb]
  \centering
  \setlength{\tabcolsep}{0pt}
  \renewcommand{\arraystretch}{1.0}
  \fontsize{8.1pt}{9pt}\selectfont
  \begin{tabular}{@{}c@{\hspace{0.016\columnwidth}}c@{}}
    \multicolumn{2}{c}{\textbf{(a) Wrist-camera invisible}} \\[1pt]
    \includegraphics[width=0.392\columnwidth]{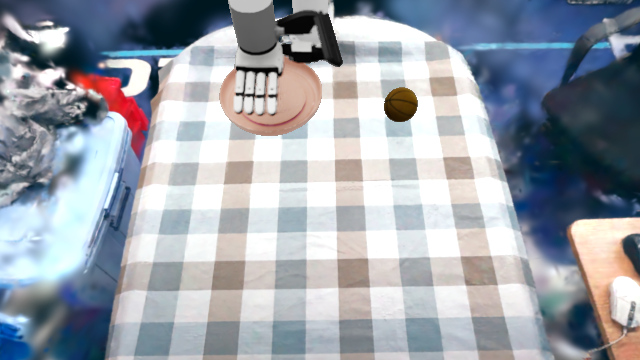} &
    \includegraphics[width=0.392\columnwidth]{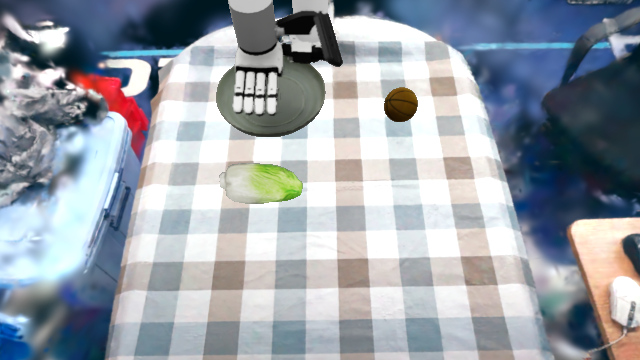} \\[-1pt]
    {Control} & {Experimental} \\[1pt]
    \multicolumn{2}{c}{\textbf{(b) Similar-object distractors}} \\[1pt]
    \includegraphics[width=0.392\columnwidth]{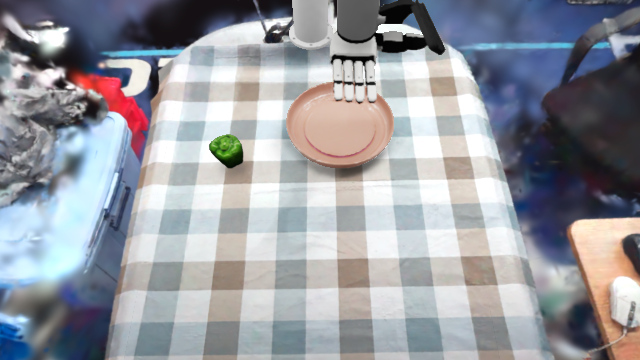} &
    \includegraphics[width=0.392\columnwidth]{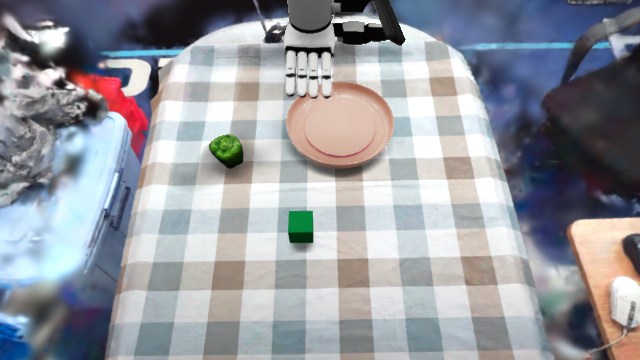} \\[-1pt]
    {Control} & {Experimental} \\[1pt]
    \multicolumn{2}{c}{\textbf{(c) Pass-by interaction}} \\[1pt]
    \includegraphics[width=0.392\columnwidth]{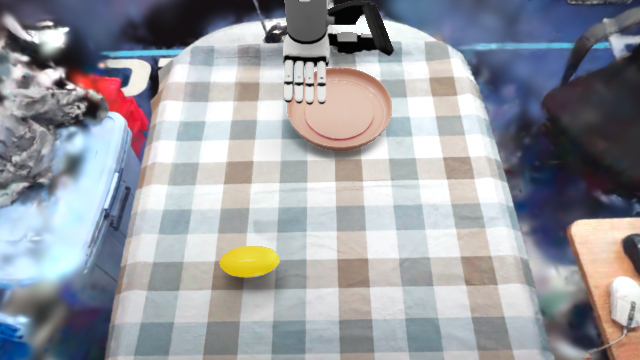} &
    \includegraphics[width=0.392\columnwidth]{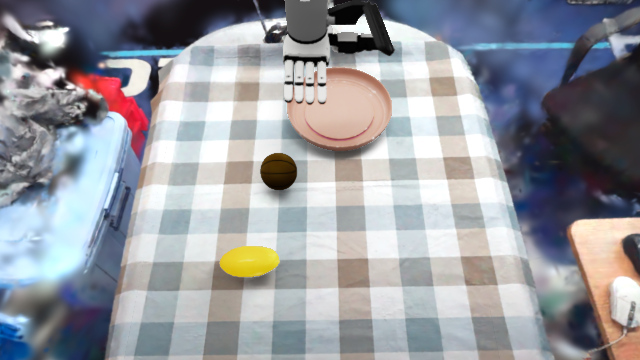} \\[-1pt]
    {Control} & {Experimental} \\[1pt]
    \multicolumn{2}{c}{\textbf{(d) Richer semantics}} \\[1pt]
    \includegraphics[width=0.392\columnwidth]{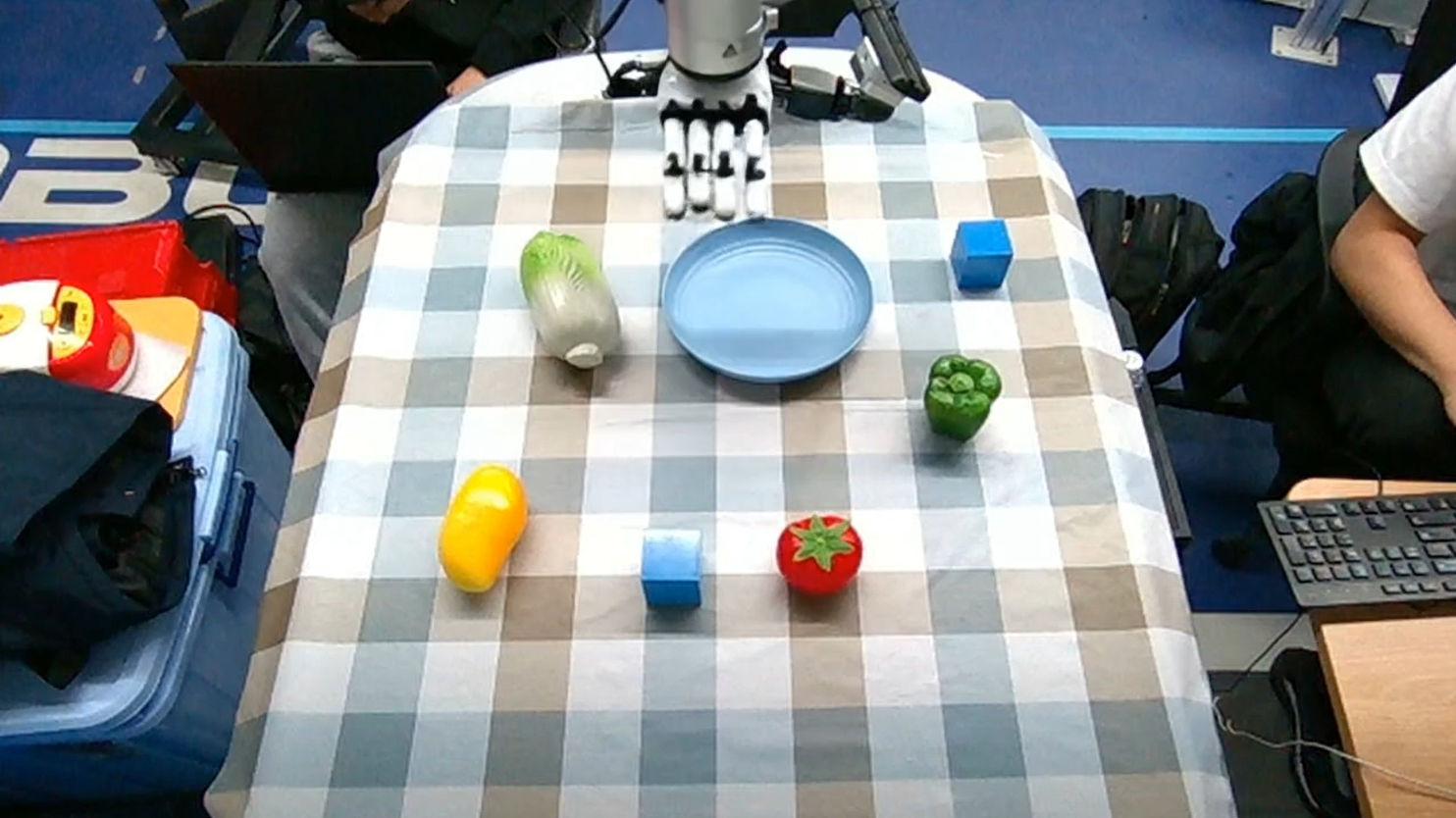} &
    \includegraphics[width=0.392\columnwidth]{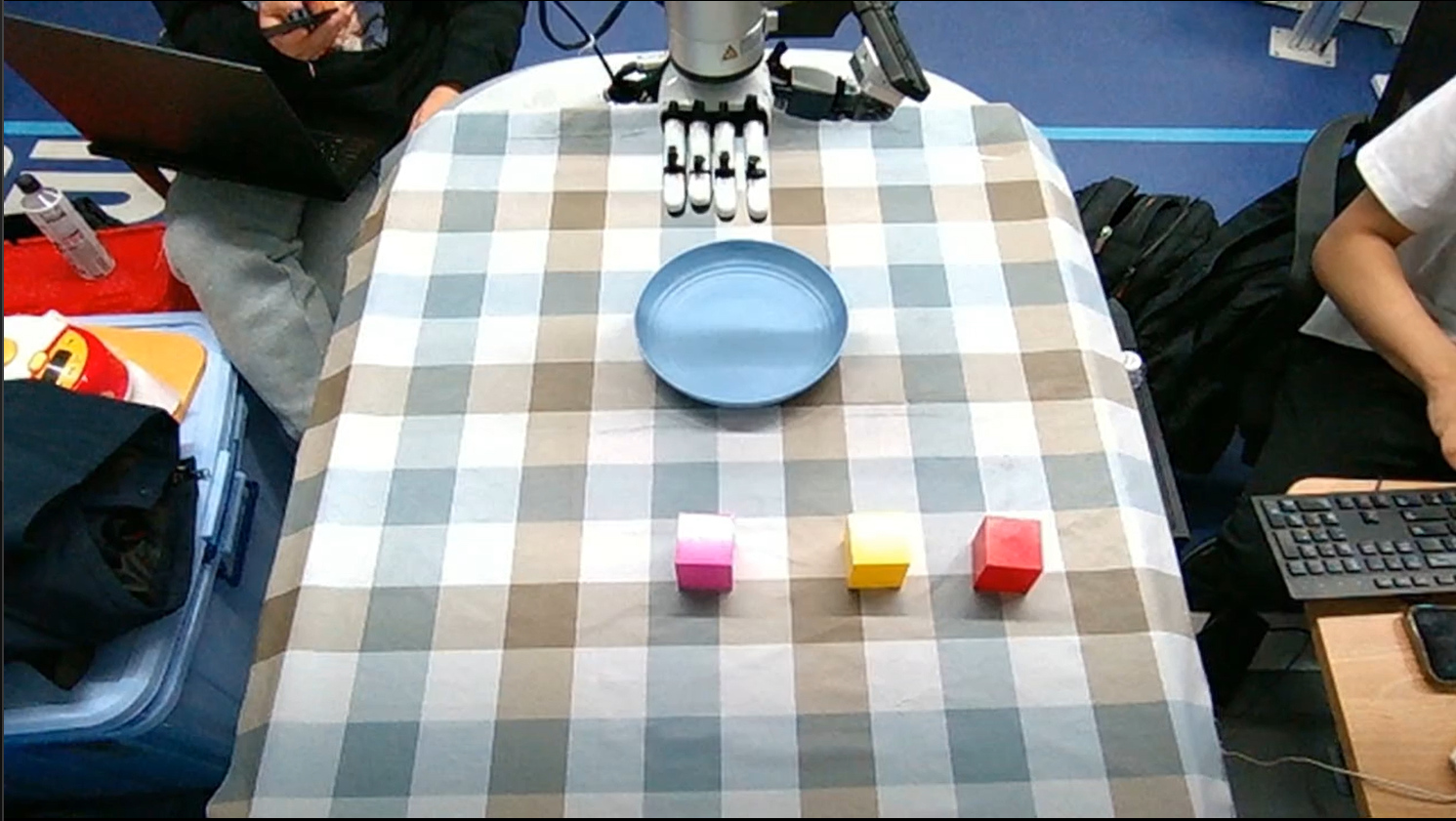} \\[-1pt]
    {Blue cube closest to tomato} & {Cube in the middle}
  \end{tabular}%
  \caption{Four evaluation settings. Rows (a)--(c) pair simulation Control scenes (left) with Experimental scenes containing a distractor (right). Row (d) shows two real robot scenes requiring relational target descriptions: proximity to another object (left) and relative ordering (right).}
  \label{fig:setting_overview}
\end{figure}

\paragraph{Simulation settings}
We construct three simulation settings where the VLA baseline is prone to errors: 
(1) \textbf{Wrist-camera invisible.} The target lies outside the wrist-camera view while a different non-target object remains visible; Control contains only the target. Object identities are randomly sampled across trials. Tested only in simulation, this setting evaluates reaching the instructed target despite the distractor's wrist-view visibility.
(2) \textbf{Similar-object distractors.} Two nearby objects of similar color and shape are visible to the wrist camera, increasing ambiguity; Control contains only the target. Mango/potato, mango/yellow cube, green pepper/green cube, and tomato/red cube pairs test fine-grained target discrimination.
(3) \textbf{Pass-by interaction.} A dissimilar distractor along the reaching path to the target may induce unintended contacts or attention shifts; Control removes it. Both objects are visible in the wrist view, with identities randomly sampled across trials. Each simulation setting uses 30 trials per condition; Fig.~\ref{fig:setting_overview}(a)--(c) shows the corresponding Control and Experimental scenes.

\paragraph{Real robot settings}
We test three real robot settings:
(1) \textbf{Similar-object distractors.} We use visually similar objects as in simulation (16 trials).
(2) \textbf{Pass-by interaction.} We place a distractor along the reaching path as in simulation (13 trials).
(3) \textbf{Richer semantics.} More complex scenes and compositional instructions require higher-level semantic grounding, such as selecting the only fruit among multiple objects or an instance satisfying a relational constraint (19 trials). Figure~\ref{fig:setting_overview}(d) shows two examples: selecting the blue cube closest to the tomato and the cube in the middle. Both require resolving spatial relations among objects beyond object-name recognition.

\paragraph{Task instructions and video prompts}
The VLA template is ``pick the \{target object name or description\} and place it into the \{container name\}.'' Braced variables are instantiated per trial. Table~\ref{tab:all_results} uses full video prompts specifying the target, container, and constraints on distractor handling, scene stability, and motion continuity. To test whether gains require these constraints, we also use neutral prompts without explicit distractor discrimination instructions, while retaining general motion, camera, and hand appearance constraints; with Seedance~2.5 in the similar-object setting, they match or exceed full prompts, indicating that this comparison does not require the additional constraints.

\subsection{Evaluation Metrics}
\label{sec:metrics}

We report two complementary metrics.

\paragraph{Instruction following success}
Instruction following success (IF) requires the shortest Euclidean distance from the end-effector position to the target surface to fall within the specified reaching threshold before any wrong object is reached. Reaching a wrong object first immediately fails IF, even if the target is reached later; every IF failure also fails the overall task.

\paragraph{Overall task success}
Overall success requires IF success and a time during the rollout when the target object's position is within the placement threshold of the container reference point (Euclidean distance) and its speed is within the staticness threshold.

\subsection{Evaluation Results}
\label{sec:exp_results}

\begin{figure}[!htb]
  \centering
  \includegraphics[width=0.9\columnwidth]{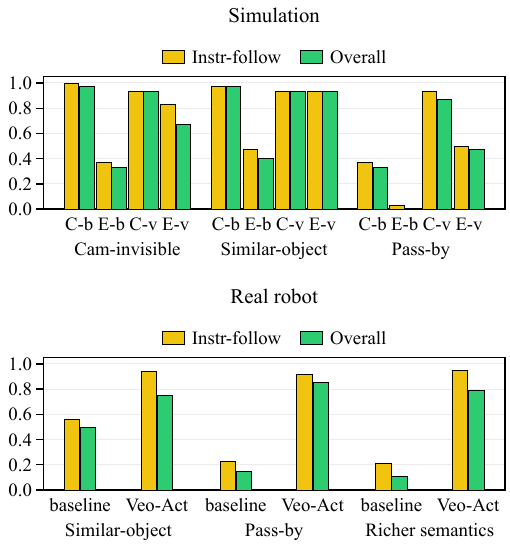}
  \caption{Simulation (top) and real robot (bottom) success rates. Yellow denotes instruction following; green denotes overall success. Simulation labels combine C/E for Control/Experimental with b/v for Baseline/Veo-Act.}
  \label{fig:sim_bar}
  \label{fig:real_bar}
\end{figure}

Table~\ref{tab:all_results} and Fig.~\ref{fig:sim_bar} report results with \(\pi_{0.5}\)~\cite{intelligence2025pi_} as both baseline and Veo-Act's low-level policy, and VPP~\cite{hu2024video} as an additional simulation baseline. The main experiments test whether a single handoff suffices to demonstrate the hierarchy's benefit; after handoff, the VLA remains active until termination. Suc/All and Rate denote successful/total trials and their fraction, respectively.

\noindent\textbf{Simulation.}
Confounds reduce baseline IF and overall success by 51.7--90.9\% and 58.6--100.0\%, respectively, relative to Control, using success counts before rounding. Under Control, Veo-Act has slightly lower observed success on wrist-camera invisible and similar-object distractors, but improves pass-by interaction from 0.37/0.33 to 0.93/0.87 (IF/Overall).

Under Experimental, Veo-Act improves both metrics in all three settings. Pooled overall success rises from the baseline's 22/90 to 62/90 (2.8\(\times\)); across Control and Experimental, it rises from 90/180 to 144/180 (+60.0\%).

\noindent\textbf{Real Robot.}
Veo-Act improves both metrics in all three real robot settings, raising pooled overall success from 12/48 to 38/48 (3.2\(\times\)). Across all simulated and real robot trials, overall success rises from 102/228 (45\%) to 182/228 (80\%), a +78.4\% relative improvement.

\newcommand{\thincline}[1]{\noalign{\global\arrayrulewidth=0.36pt}\cline{#1}\noalign{\global\arrayrulewidth=0.675pt}}
\subsubsection{Ablation Study}

Tables~\ref{tab:pure_idm_ablation}, \ref{tab:ablation_study}, and~\ref{tab:video_backbones} report execution strategy, IDM architecture, and video model ablations, respectively.
Architecture and video model ablations use simulation Experimental settings with 30 trials per setting and variant. Tables~\ref{tab:ablation_study} and~\ref{tab:video_backbones} bold the best rates per metric and setting, including ties; Table~\ref{tab:ablation_study} shades the full model.

\paragraph{\textbf{Execution Strategy Ablation}}
To test direct low-level control from generated video in simulation, Pure IDM executes the inferred, smoothed action chunk throughout the episode without switching to the VLA.

\begin{table}[!htb]

  \centering
  \caption{\textbf{Pure IDM execution results.}}
  \label{tab:pure_idm_ablation}
  \vspace{0.2em}
  \tablebodystyle
  \setlength{\tabcolsep}{0.9pt}

  \begin{tabular*}{\linewidth}{@{\extracolsep{\fill}}cccccc@{}}
    \hline
    \multirow{2}{*}{\textbf{Task}}
    & \multirow{2}{*}{\textbf{Setting}}
    & \multicolumn{2}{c}{\textbf{Instruction following}}
    & \multicolumn{2}{c}{\textbf{Overall}} \\
    \thincline{3-6}
    & & \textbf{Suc/All} & \textbf{Rate} & \textbf{Suc/All} & \textbf{Rate} \\
    \hline
    \multirow{2}{*}{\begin{tabular}[c]{@{}c@{}}Wrist-camera\\invisible\end{tabular}}
    & Control      & 23/30 & 0.77 & 2/30 & 0.07 \\
    & Experimental & 14/30 & 0.47 & 1/30 & 0.03 \\
    \hline
    \multirow{2}{*}{\begin{tabular}[c]{@{}c@{}}Similar-object\\distractors\end{tabular}}
    & Control      & 17/30 & 0.57 & 1/30 & 0.03 \\
    & Experimental & 10/30 & 0.33 & 0/30 & 0.00 \\
    \hline
    \multirow{2}{*}{\begin{tabular}[c]{@{}c@{}}Pass-by\\interaction\end{tabular}}
    & Control      & 7/30  & 0.23 & 0/30 & 0.00 \\
    & Experimental & 5/30  & 0.17 & 0/30 & 0.00 \\
    \hline
  \end{tabular*}

\end{table}

Pure IDM sometimes follows instructions, especially with wrist-camera invisibility, but has very low overall success across all tasks. Grasping is a major bottleneck: the robot often reaches the target but fails to grasp stably and complete the task. Generated hand deformation and physically implausible contacts can impair inferred manipulation actions. These results motivate hierarchical execution.

\paragraph{\textbf{IDM Architecture Ablation}}
We next ablate the IDM architecture and training choices on the wrist-camera invisible setting, including a ResNet backbone~\cite{he2016deep}, no noise augmentation, and separately trained single-head models. The latter variant trains separate action and interaction gate predictors without a shared encoder or joint objective; the controller still switches using the learned gate. These variants achieve lower overall success than the full model. In particular, the separate-model variant matches its IF success (25/30) but has lower overall success (17/30 versus 20/30). This supports the joint architecture for task completion but does not isolate contributions from action prediction, gate prediction, or their combination.

\begin{table}[!htb]

    \centering
    \caption{\textbf{IDM architecture ablation results.}}
    \label{tab:ablation_study}
    \vspace{0.2em}
    \tablebodystyle
    \setlength{\tabcolsep}{1.35pt}

    \begin{tabularx}{\linewidth}{@{}l*{4}{>{\centering\arraybackslash}X}@{}}
      \hline
      & \multicolumn{4}{c}{\textbf{Wrist-camera invisible}} \\
      \thincline{2-5}
      \textbf{Variant}
      & \multicolumn{2}{c}{\textbf{Instruction following}}
      & \multicolumn{2}{c}{\textbf{Overall}} \\
      \thincline{2-5}
      & \textbf{Suc/All} & \textbf{Rate}
      & \textbf{Suc/All} & \textbf{Rate} \\
      \hline
      ResNet backbone
      & 22/30 & 0.73
      & 17/30 & 0.57 \\
      Without noise
      & 22/30 & 0.73
      & 16/30 & 0.53 \\
      Separate models
      & 25/30 & \textbf{0.83}
      & 17/30 & 0.57 \\
      \rowcolor{gray!35}
      Ours
      & 25/30 & \textbf{0.83}
      & 20/30 & \textbf{0.67} \\
      \hline
    \end{tabularx}

\end{table}

\begingroup
\setlength{\intextsep}{6pt}
\paragraph{\textbf{Video Model Ablation and Baseline Comparison}}
We test the hierarchy's dependence on the video generator by replacing Veo-3.1-fast with Seedance~2.5 in the pass-by interaction and similar-object distractor settings. The two variants share the same IDM and VLA, switching configuration, video prompt, and initial scene configurations. Table~\ref{tab:video_backbones} also includes $\pi_{0.5}$ and Fast-WAM~\cite{yuan2026fastwam} as separate baselines. Fast-WAM is trained on the same robot data as $\pi_{0.5}$. Although its pass-by instruction following rate improves, none of its evaluated rollouts completes a grasp, and its overall success is zero in both settings. In contrast, Veo-3.1-fast and Seedance~2.5 yield comparable success rates within Veo-Act on these two tasks. The hierarchy thus extends beyond Veo-3.1-fast in the tested settings.

\begin{table}[!htb]
  \centering
  \caption{\textbf{Video model ablation and baseline comparisons.}}
  \label{tab:video_backbones}
  \tablebodystyle
  \setlength{\tabcolsep}{1.35pt}
  \begin{tabular*}{\linewidth}{@{\extracolsep{\fill}}lcccc@{}}
    \hline
    & \multicolumn{2}{c}{\textbf{Pass-by}} & \multicolumn{2}{c}{\textbf{Similar-object}} \\
    \thincline{2-3}\thincline{4-5}
    \textbf{Method} & \textbf{IF} & \textbf{Overall} & \textbf{IF} & \textbf{Overall} \\
    \hline
    $\pi_{0.5}$ & 0.03 & 0.00 & 0.47 & 0.40 \\
    Fast-WAM & 0.20 & 0.00 & 0.47 & 0.00 \\
    \hline
    Veo-Act (Veo-3.1-fast) & \textbf{0.50} & 0.47 & \textbf{0.93} & \textbf{0.93} \\
    Veo-Act (Seedance~2.5) & \textbf{0.50} & \textbf{0.50} & 0.90 & 0.90 \\
    \hline
  \end{tabular*}
\end{table}

\subsubsection{Failure Analysis}
\label{sec:failure_analysis}
We analyze the 28 failures among 90 Veo-Act rollouts in the main evaluation's three simulation Experimental settings. We identify video generation failures when the visual plan selects the wrong target or contains severe artifacts. Among the remaining execution failures, we distinguish guidance (non-interaction) from interaction using the phase definitions from IDM training. We observe 21 video generation failures (75.0\%), 1 guidance failure (3.6\%), and 6 interaction failures (21.4\%).

Inspection of the 21 failed Veo-Act videos identifies 8 cases of hallucinated objects (38.1\%), 8 cases of generating or selecting the wrong object (38.1\%), and 7 cases of severe hand morphing artifacts (33.3\%). A generated video may belong to multiple error subcategories; all percentages use these 21 failures as the denominator. These observations document hand deformation in failed generated plans without measuring its prevalence across all generated videos or independently isolating its causal contribution to execution failure.

\subsection{Runtime and Deployment}
\label{sec:runtime}

Table~\ref{tab:runtime} separates one-time planning from online execution costs; dashes mark planning stages absent from the baseline. Video generation and IDM conversion take 48.7\,s and 9.97\,s before execution. Veo-Act then executes the plan and reactive VLA without further video queries at 379\,ms per step (2.64\,Hz), versus 324\,ms (3.09\,Hz) for $\pi_{0.5}$, adding startup latency and online overhead. These system-level measurements do not imply equal compute: the proprietary video model's serving compute is unavailable.

\begin{table}[!htb]
  \centering
  \caption{\textbf{Runtime and deployment costs.}}
  \label{tab:runtime}
  \tablebodystyle
  \setlength{\tabcolsep}{5.4pt}
  \begin{tabular*}{\linewidth}{@{\extracolsep{\fill}}lcc@{}}
    \hline
    \textbf{Measurement} & $\boldsymbol{\pi_{0.5}}$ & \textbf{Veo-Act} \\
    \hline
    \multicolumn{3}{l}{\textit{One-time planning}} \\
    Video generation (s) & -- & 48.7 \\
    IDM conversion (s) & -- & 9.97 \\
    \hline
    \multicolumn{3}{l}{\textit{Online execution}} \\
    RPC latency (ms) & 270 & 325 \\
    Step time (ms) & 324 & 379 \\
    Execution frequency (Hz) & 3.09 & 2.64 \\
    \hline
    GPU memory (MB) & 844 & 1487 \\
    \hline
  \end{tabular*}
\end{table}

\endgroup

\section{Limitations}
Veo-Act remains sensitive to video fidelity: changes in robot appearance, viewpoint, or scene layout can alter predictions and downstream planning. Video generation failures dominate (Sec.~\ref{sec:failure_analysis}), motivating more reliable plans.

Interaction can fail because of imperfect approach configurations or reactive control errors, motivating better handoffs and interaction control. Such failures should not all be attributed to generated hand deformation.

Long-horizon tasks and tasks requiring higher dexterity remain to be evaluated.

\section{Conclusion}
\label{sec:conclu}

Veo-Act uses video model priors to guide motion toward the target and delegates precise interaction to a reactive VLA. A single learned handoff improves success in the evaluated confounded settings, supporting this division of labor. Future work will improve plan reliability and handoff accuracy and broaden task coverage.

\section*{ACKNOWLEDGMENT}
GPT-Image generated some icons in Figs.~\ref{fig:intro_teaser}, \ref{fig:pipeline}, and~\ref{fig:train}.

\FloatBarrier

\clearpage

\makeatletter
\let\arxiv@ruled\fs@ruled
\def\fs@ruled{\arxiv@ruled\def\@fs@pre{\vskip5pt\hrule height.8pt depth0pt\kern2pt}}
\restylefloat{algorithm}
\makeatother
\useRomanappendicesfalse
\appendices

\section{IDM Representation and Supervision}
\label{app:representation}

This appendix details the implementation of the visual transition model and its interface with the controller. We retain the notation of Sec.~\ref{sec:method}: $i$ indexes recorded training samples, $j$ indexes transitions in the generated video, $k$ indexes the smoothed queue, and $t$ indexes real execution steps.

\subsection{Visual Features and Prediction Heads}

Each of the two global images is processed by a DINOv3 ViT-B/16 encoder. Spatial averaging of its patch features produces a vector of 768 dimensions. Concatenating the two frame features yields a vector of 1536 dimensions, which is supplied to both prediction heads. Each head uses a linear layer with 256 outputs, a ReLU activation, and a final linear projection. The action projection has 21 outputs; the interaction projection has one output, converted to a probability by a sigmoid. Table~\ref{tab:app_representation} summarizes this interface. The visual features are shared between the two prediction tasks, and the encoders are updated during IDM training.

\begin{table}[H]
\centering
\caption{\textbf{IDM features and outputs.}}
\label{tab:app_representation}
\tablebodystyle
\begin{tabular*}{\linewidth}{@{\extracolsep{\fill}}lr@{}}
\hline
\textbf{Component} & \textbf{Dimensions} \\
\hline
Pooled visual feature per frame & 768 \\
Concatenated features from two frames & 1536 \\
Hidden layer in each head & 256 \\
End effector position & 3 \\
End effector orientation representation & 6 \\
Hand joint configuration & 12 \\
Interaction probability & 1 \\
\hline
\end{tabular*}
\end{table}

The action target describes the configuration at the later frame of a recorded pair. It is an absolute target rather than a frame displacement or a velocity. The six orientation coordinates represent three rotational degrees of freedom; they are not six additional robot joints. Together with position, they specify the end effector pose used by inverse kinematics for the seven arm joints. The twelve hand coordinates directly describe the hand configuration. Thus, the 21 action coordinates and the robot's 19 joint degrees of freedom describe different representations of the same control system. The interaction probability is separate from both counts.

\subsection{Training Samples and Loss Assignment}

Table~\ref{tab:app_supervision} separates samples according to their available supervision. The 300k simulation pairs include binary phase annotations: 1 denotes interaction and 0 denotes non-interaction. Both positive and negative annotated examples supervise the gate. The additional 100k simulation pairs and 150k real pairs supervise action prediction only. In total, the action model uses 550k frame pairs.

\begin{table}[H]
\centering
\vspace*{6pt}
\caption{\textbf{Supervision available for each data source.}}
\label{tab:app_supervision}
\tablebodystyle
\begin{tabular*}{\linewidth}{@{\extracolsep{\fill}}lrrr@{}}
\hline
\textbf{Source} & \textbf{Pairs} & \textbf{Action} & \textbf{Gate} \\
\hline
Simulation with phase annotations & 300k & Yes & Yes \\
Additional random simulation motion & 100k & Yes & No \\
Real robot motion & 150k & Yes & No \\
\hline
\end{tabular*}
\end{table}

Non-interaction includes navigation and transport after interaction completion. The label describes the execution phase, rather than whether an object is currently held. This distinction matters when an object remains in the hand during subsequent transport. Action regression uses Huber loss, while phase prediction uses binary cross entropy. A sample without a phase annotation contributes no gate loss. Robot configurations and phase annotations are training targets; neither is appended to the IDM's visual input.

Joint training refers specifically to optimization of the IDM encoders and its two heads. The video generator and VLA are separate components. The architecture ablation with separate models instead trains an action predictor and a gate predictor independently. It retains the switching controller, so its comparison with the shared model concerns joint representation learning for task completion.

\section{Action Queue Construction and Execution}
\label{app:execution}

\subsection{Smoothing and Temporal Alignment}

The raw plan stores an action vector and its predicted interaction probability in each row. These quantities are concatenated before smoothing. For each channel, the smoother selects endpoints and local extrema by recursively splitting sufficiently long segments. It applies a centered moving average within the resulting intervals, retaining their boundary values. This reduces local variation while preserving selected changes in the trajectory.

\begin{figure}[!htb]
\centering
\includegraphics[width=0.98\columnwidth]{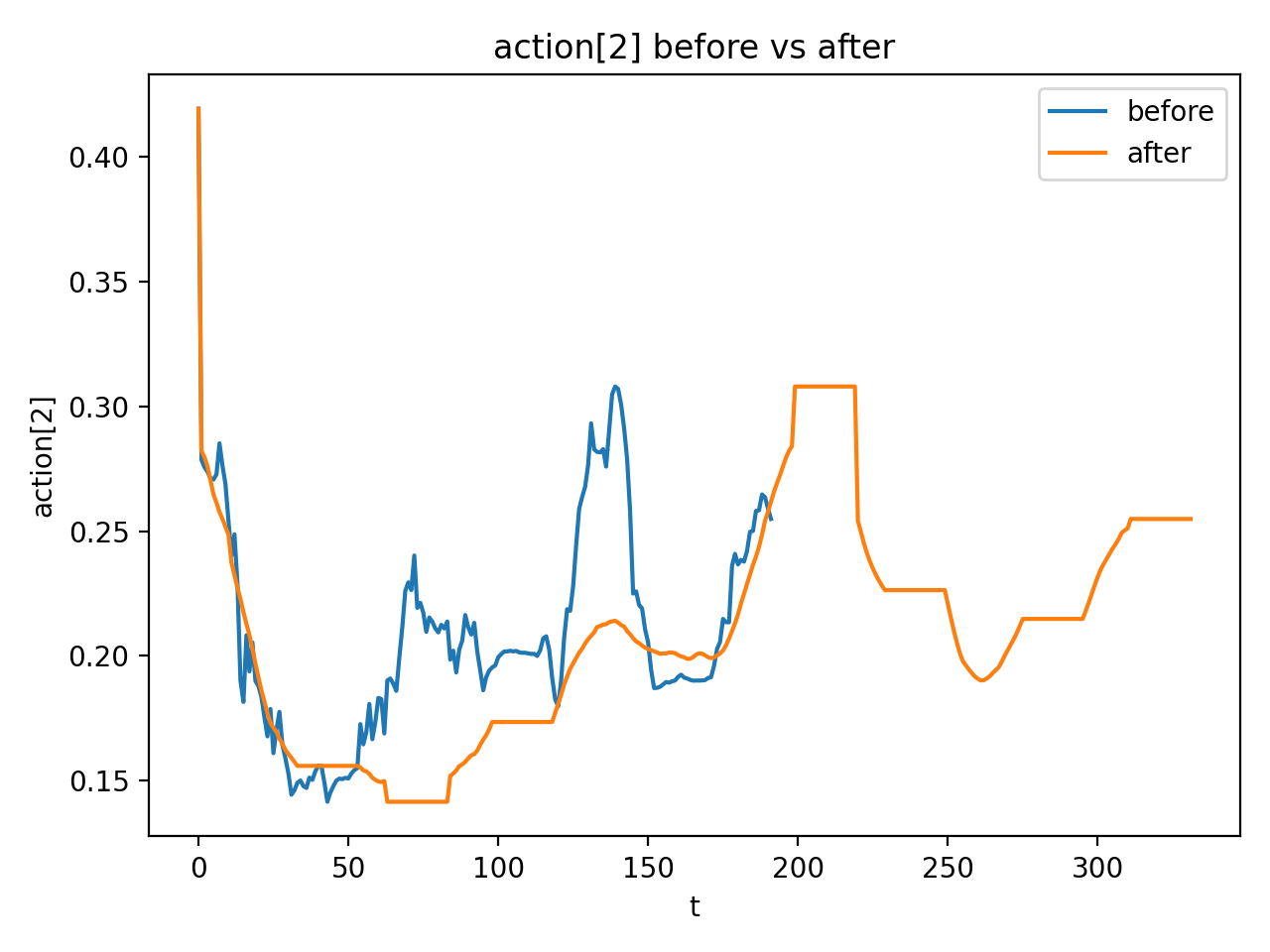}
\caption{\textbf{Illustration of action smoothing.} One action coordinate before and after smoothing and temporal extension. The plot's horizontal label $t$ denotes the local sample index in each curve, rather than a shared execution clock; repeating entries changes the sequence length.}
\label{fig:app_smoother}
\end{figure}

Temporal extension then repeats complete rows at selected keypoints. Consequently, every inserted action has a gate value at the same queue index. The smoother does not independently stretch the action sequence and the gate sequence. This preserves their correspondence even when the resulting queue contains more entries than the generated video has transitions. Figure~\ref{fig:app_smoother} illustrates the transformation for one action coordinate.

\subsection{Planned Gates and Observed Gates}

There are two uses of the interaction head. Applying it to generated frame pairs gives the planned gate sequence $G_j^{*}$. Applying it to consecutive real images gives the online probability $G_t$. The planned gates describe phases within the video plan; the online gate determines when the robot should hand control to the VLA. These sequences therefore need not cross their thresholds at the same numerical index.

The main experiments use a single handoff, summarized in Algorithm~\ref{alg:app_execution}. A persistence window requires the online gate to remain above its threshold before enabling the VLA. Once enabled, the VLA receives the current global image, wrist image, robot state, and task instruction at each step and remains active until termination. The fixed video plan is prepared before execution. Evaluating the gate online does not invoke the video generator again.

\begin{algorithm}[!htb]
\caption{Execution with a single VLA handoff}
\label{alg:app_execution}
\begin{algorithmic}[1]
\Require Initial global image $I_0$, instruction $\ell$
\Require Gate threshold $\tau$, persistence length $p$
\State Generate video $I^{*}_{0:n}$ from $I_0$ and $\ell$
\For{$j=0,\ldots,n-1$}
  \State $(a_j^{*},G_j^{*})\gets f_{\mathrm{IDM}}(I_j^{*},I_{j+1}^{*})$
\EndFor
\State Smooth paired rows into $\mathcal Q=\{(\hat a_k^{*},\hat G_k^{*})\}_{k=0}^{m}$
\State $t\gets0$; $k\gets0$; $c\gets0$; $\textit{useVLA}\gets\textbf{false}$
\While{episode has not terminated}
  \State Observe $I_t,W_t,s_t$
  \If{a previous real image is available}
    \State $G_t\gets\operatorname{gate}(f_{\mathrm{IDM}}(I_{t-1},I_t))$
    \If{$G_t>\tau$}
      \State $c\gets c+1$
    \Else
      \State $c\gets0$
    \EndIf
    \If{$c\ge p$}
      \State $\textit{useVLA}\gets\textbf{true}$
    \EndIf
  \EndIf
  \If{$\textit{useVLA}$}
    \State $a_t\sim\pi_{\mathrm{VLA}}(\cdot\mid s_t,(I_t,W_t),\ell)$
  \ElsIf{$k\le m$}
    \State $a_t\gets\hat a_k^{*}$; $k\gets k+1$
  \Else
    \State $a_t\gets$ current robot configuration
  \EndIf
  \State Execute $a_t$, retain $I_t$, and set $t\gets t+1$
\EndWhile
\end{algorithmic}
\end{algorithm}

The general controller in Sec.~\ref{sec:method} also permits a return to the remaining queue. During VLA execution, its cursor is paused. On return, the controller skips consecutive stored entries whose planned gate exceeds the queue threshold, then resumes at the first remaining low gate entry. This operation follows queue order rather than matching timestamps between the real rollout and generated video. The main quantitative results use the single handoff configuration described above.

\begin{figure*}[!t]
\centering
\subfloat[Blue cube closest to the tomato]{\includegraphics[width=0.31\textwidth]{figures/image_02.jpg}}
\hfill
\subfloat[Fruit farthest from the plate]{\includegraphics[width=0.31\textwidth]{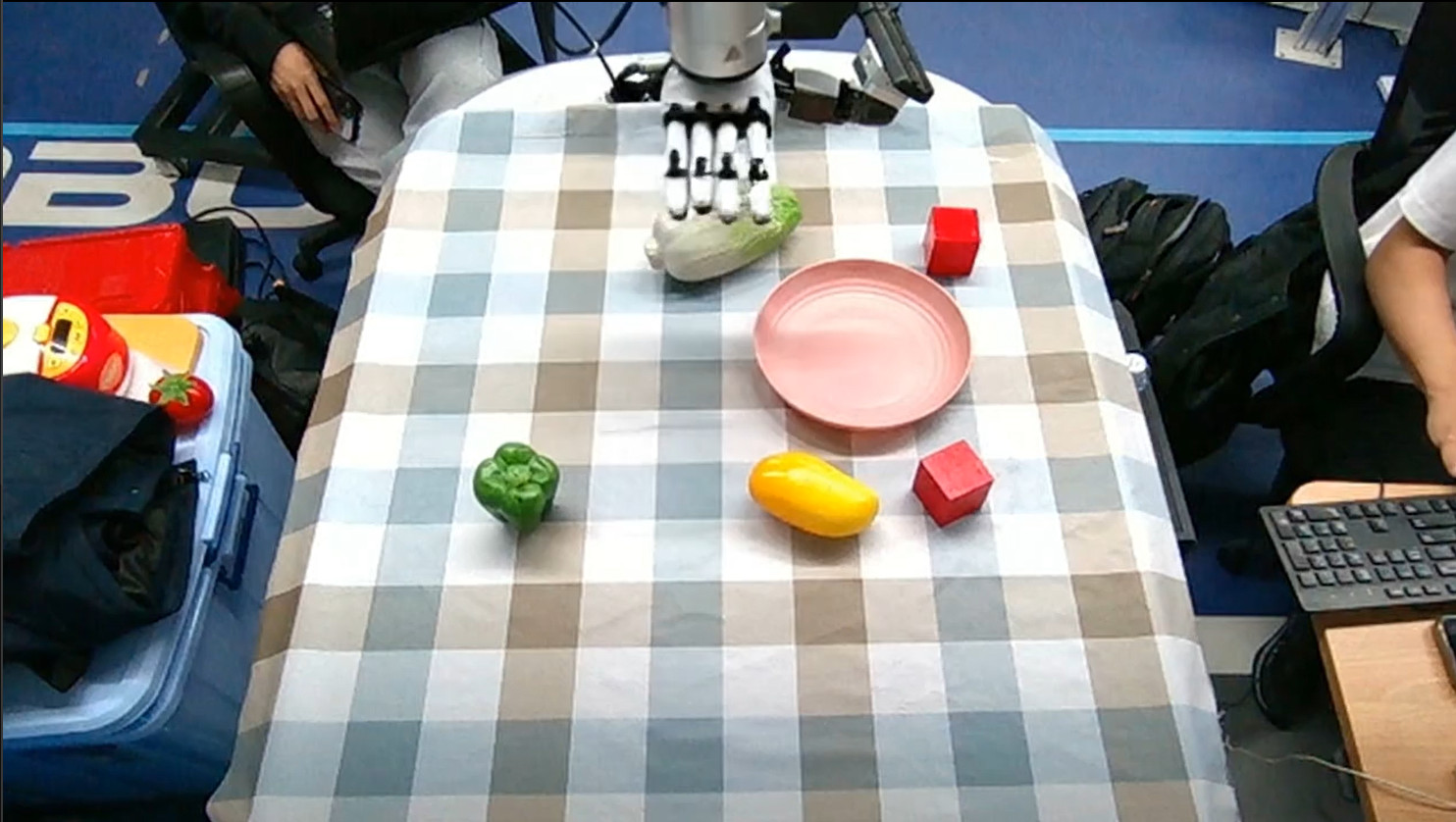}}
\hfill
\subfloat[Cube in the middle]{\includegraphics[width=0.31\textwidth]{figures/image_04.jpg}}
\caption{\textbf{Examples of relational target descriptions.} The scenes illustrate proximity to a reference object, category selection followed by a distance comparison, and relative ordering. Panels (a) and (c) enlarge examples shown in the main text; panel (b) adds a complementary scene.}
\label{fig:app_semantic}
\end{figure*}

\section{Evaluation Protocol and Interpretation}
\label{app:protocol}

\subsection{Scene Conditions and Observation Access}

Control and Experimental refer to scene conditions, not to different methods. Within each simulation pair, the instructed target starts at the same position. Experimental introduces a distractor configuration; Control removes the distractor. Comparing methods within each condition tests their behavior under the same type of visual ambiguity, while comparing conditions measures the effect of the confound. Each simulation setting has 30 trials per condition and method.

The three simulation settings probe distinct aspects of target selection. The wrist visibility setting places the target outside the wrist camera view while leaving a distractor visible. Similar objects introduce ambiguity in appearance: the tested pairs include mango and potato, mango and yellow cube, green pepper and green cube, and tomato and red cube. The pass-by setting places a visually dissimilar distractor along the route to the target. In this case the challenge is to continue toward the instructed object despite an intervening object.

Both the baseline VLA and Veo-Act's VLA receive the global and wrist images, robot state, and language instruction. The video model and IDM use the global view. The comparison therefore does not give Veo-Act an extra camera view unavailable to the baseline. The OOD designation refers to objects, scene configurations, and tasks absent from the IDM and VLA robot training data.

\subsection{Relationship Between the Two Success Metrics}

Instruction following measures whether the robot reaches the instructed target before any wrong object. A rollout that first reaches a distractor fails IF even if it subsequently reaches the target. Since overall success requires IF success, such a rollout also fails the task. Conversely, reaching the correct target does not by itself establish overall success: the object must also satisfy the placement and staticness criteria in Sec.~\ref{sec:metrics}.

This distinction separates target selection from task completion. For example, a rollout may reach the correct object but fail to complete its transfer into the container. It contributes to IF success but not overall success. The two columns in the result tables therefore describe related, nested events rather than independent scores. Suc/All reports the count of successes and its evaluation denominator; Rate is the corresponding fraction rounded for display. Aggregated results sum success counts and trial counts before computing the rate.

\subsection{Failure Categories}

The failure analysis concerns the 28 unsuccessful Veo-Act rollouts among the 90 simulation Experimental trials. Videos that select the wrong target or contain severe artifacts are assigned to video generation failure. The remaining execution failures are classified by the guidance and interaction phase definitions used for IDM training. This yields 21 video generation failures, one guidance failure, and six interaction failures, as reported in the main text.

Within the 21 failed videos, object hallucination, wrong object generation or selection, and severe hand deformation are descriptive tags. A single video can receive several tags; their counts of 8, 8, and 7 must therefore not be added to estimate the number of failed rollouts. These observations explain the inspected failures and are not a separate estimate of artifact frequency over all generated videos. They preserve the distinction between a prior that motivates visual planning and the reliability of an individual generated plan.

\section{Language Instructions and Semantic Examples}
\label{app:language}

\subsection{Prompt Instantiation}

The VLA template is ``pick the \{target object name or description\} and place it into the \{container name\}.'' Fields are instantiated per trial. Relational phrases occupy the same target field as object names, without supplying a separately resolved target identity.

The full prompt below, reproduced from the existing specification, names the target and container and constrains scene stability, distractor handling, and motion continuity. Variables accept object names or relational descriptions. The neutral variant removes instructions to distinguish similar objects and leave other objects undisturbed, while retaining smooth motion, a fixed camera, and consistent hand appearance. Thus, neutral means no explicit distractor handling instructions; it does not mean unconstrained generation.

\begin{tcolorbox}[colback=gray!5,colframe=gray!55,boxrule=0.4pt,arc=0pt,
left=1.5mm,right=1.5mm,top=1.2mm,bottom=1.2mm,
fontupper=\small,title={Full video prompt}]
A realistic video showing a robot with a human-like dexterous hand.
In front of the robot, there is an existing \{target\_object\} to be grasped and a \{container\} on the table.
The robot should clearly distinguish the \{target\_object\} from other similar-looking objects on the table and must not confuse them.
Any object on the table other than the \{target\_object\} should remain undisturbed throughout the entire process.
The \{target\_object\} should remain in its original position until it is picked up by the robot.
The robot gently reaches out its human-like hand, picks up the \{target\_object\}, and carefully places it into the \{container\} without pausing or lingering midway.
The scene should maintain the same viewpoint as the reference image throughout the entire video, without any change in camera angle or camera position.
The movement should be smooth and natural, showing clear human-like hand motion and physical interaction with the \{target\_object\}.
The \{container\} should not be moved during the process.
\end{tcolorbox}

The neutral template below reproduces the single arm implementation and uses the same variable substitution.

\begin{tcolorbox}[colback=gray!5,colframe=gray!55,boxrule=0.4pt,arc=0pt,
left=1.5mm,right=1.5mm,top=1.2mm,bottom=1.2mm,
fontupper=\small,title={Neutral video prompt}]
A realistic video showing a robot with a human-like dexterous hand performing a pick-and-place task.
The robot smoothly reaches for the \{target\_object\}, grasps it, and places it into the \{container\}.
The motion should be smooth, natural, and continuous, with realistic physical interaction.
The robot hand and fingers should remain consistent with the reference image throughout the video, without deformation, warping, or changes in shape.
The camera should remain fixed at the same viewpoint as the reference image.
\end{tcolorbox}

\subsection{Resolving Relational Descriptions}

Figure~\ref{fig:app_semantic} adds a third semantic scene from the original supplementary material. These illustrations do not add trials to the 19 real robot evaluations in Table~\ref{tab:all_results}.

The blue cube closest to the tomato requires identifying the reference object and comparing candidate cube locations. The fruit farthest from the plate requires selecting the fruit category before comparing distances to the plate. The cube in the middle requires resolving relative ordering. Category or color alone cannot identify these targets; their relational descriptions fill the same instruction template used for simpler scenes.

\end{document}